\documentclass[letterpaper,conference]{IEEEtran}
\usepackage[letterpaper,left=0.75in,right=0.75in,bottom=0.75in,top=0.75in]{geometry}
\IEEEoverridecommandlockouts
\usepackage{cite}
\usepackage{hyperref}
\usepackage{amsmath,amssymb,amsfonts}
\DeclareMathOperator*{\argmin}{arg\,min}
\usepackage{algorithmic}
\usepackage{graphicx}
\usepackage{textcomp}
\usepackage{xcolor}
\usepackage{color,soul} 
\def\BibTeX{{\rm B\kern-.05em{\sc i\kern-.025em b}\kern-.08em
    T\kern-.1667em\lower.7ex\hbox{E}\kern-.125emX}}
\begin{document}

\title{\vspace*{0.25in}Improving the Accuracy of Early Exits in Multi-Exit Architectures via Curriculum Learning}

\author{\IEEEauthorblockN{Arian Bakhtiarnia, Qi Zhang and Alexandros Iosifidis}
\IEEEauthorblockA{\textit{DIGIT, Department of Electrical and Computer Engineering, Aarhus University, Denmark}\\
\{arianbakh,qz,ai\}@ece.au.dk}
}

\maketitle

\begin{abstract}
Deploying deep learning services for time-sensitive and resource-constrained settings such as IoT using edge computing systems is a challenging task that requires dynamic adjustment of inference time. Multi-exit architectures allow deep neural networks to terminate their execution early in order to adhere to tight deadlines at the cost of accuracy. To mitigate this cost, in this paper we introduce a novel method called \textit{Multi-Exit Curriculum Learning} that utilizes curriculum learning, a training strategy for neural networks that imitates human learning by sorting the training samples based on their difficulty and gradually introducing them to the network. Experiments on CIFAR-10 and CIFAR-100 datasets and various configurations of multi-exit architectures show that our method consistently improves the accuracy of early exits compared to the standard training approach.
\end{abstract}


\section{Introduction}

Deep learning models have been successful in solving many problems in various domains of science and technology, ranging from autonomous vehicles to drug discovery \cite{dl_success}. However, a general drawback of deep neural networks is that, by definition, they are built from many layers of interconnected neurons. This results in models containing millions of parameters that need to be deployed on powerful processors due to their high computational cost. This restriction has sparked a great deal of research targeting neural network compression in recent years, thus many methods have been developed for the purpose of making deep learning models more lightweight; including pruning \cite{pruning}, quantization \cite{quantization}, regularization \cite{regularization} and knowledge distillation \cite{kd} to name a few.

The high computational cost of deep learning models becomes even more problematic in computationally restricted environments, such as mobile and IoT devices. Yet, deep learning has many use cases in such settings, including but not limited to video surveillance, voice assistants, network intrusion detection and augmented reality \cite{dl_edge}. Many of these use cases are time-sensitive and require applications to run with respect to strict time limits, for instance, in the cases of cooperative autonomous driving and augmented reality \cite{edge_time_limit}.

To enable time-sensitive Internet of Things applications, computationally expensive tasks, such as deep learning services, are sometimes offloaded from end-devices to edge servers using edge computing systems in order to decrease the overall execution time \cite{prev_offloading}. However, these systems often have a distributed and multi-tiered network architecture where the time required for the transmission of data between various devices is variable and depends on the communication channel state and the data size. This calls for novel neural network designs that can dynamically adapt their inference time to account for these variations in transmission time. Among lightweight deep learning methods, the concept of \textit{early exits}\cite{ee_survey} is a promising solution that particularly fits these settings, which is sometimes also referred to as \textit{multi-exit architectures} or \textit{auxiliary classifiers} in the literature.

In multi-exit architectures, branches composed of just a few layers of neurons are added at intermediate layers of a deep network called the \textit{backbone} network. Such branches are trained to perform the same task as the backbone network and produce an output similar to that of the final layer of the network, albeit they are inevitably less accurate. These branches can then be used to make inference time more dynamic at the cost of accuracy. For instance, when there is a strict time budget and it is suspected that the deadline will be missed if the entire network is traversed, the output of these early exit branches can be used instead. Another way of utilizing early exits for dynamic inference is to use the output of early exit branches for ``easier samples'' and only compute the output of later branches or the final output of the backbone network when the input sample is difficult. There are various methods for detecting where to exit, one of the easiest and most intuitive ones being to determine the confidence of the output of a branch. For instance, a strategy that is used for classification problems is to set a threshold on the entropy of the classification result \cite{branchynet}.

As previously mentioned, early exit branches are typically less accurate compared to the final output of the corresponding backbone network, therefore it is vital for them to be as accurate as possible to maintain the reliability of the output. Since the architecture of early exit branches is often very shallow in order to avoid introducing high additional overhead, increasing their accuracy is generally a challenging task. Phuong et al. \cite{ee_kd} recently showed that knowledge distillation-based training can be used to improve the accuracy of early exits.

In this paper, we propose a new approach for improving the accuracy of early exit branches based on curriculum learning. Curriculum learning \cite{original_cl} is a training strategy for neural networks that has been shown to improve the final accuracy of a network in certain cases. The idea behind curriculum learning is similar to how humans learn new tasks: a well-informed teacher can be used to initiate the training with the simplest material and gradually introduce more difficult subjects to the student. For neural networks however, sometimes the opposite approach of introducing the hardest subjects first, called \textit{anti-curriculum}, can be beneficial as well. To the best of our knowledge, curriculum learning has not yet been explored in the context of multi-exit architectures. 
We tested our proposed approach in 16 different scenarios involving multi-exit architectures for the problem of image classification, and found that it consistently increases the accuracy of early exits in every case. These scenarios involve two different datasets, namely CIFAR-10 and CIFAR-100 \cite{cifar}, four different backbone networks and two different branch locations for each backbone. We also show that the proposed approach works regardless of the optimization algorithm used during training\footnote{Our code is made available at \url{https://gitlab.au.dk/maleci/MultiExitCurriculumLearning}.}.

The remainder of the paper is structured as follows. Section \ref{S:RelatedWork} provides an overview of relevant literature. The proposed approach, called \textit{Multi-Exit Curriculum Learning}, is described in Section \ref{S:Methods}. Experimental results are provided in Section \ref{S:Results}. Finally, Section \ref{S:Conclusions} concludes the paper and briefly discusses future research directions.

\section{Related Work}\label{S:RelatedWork}
In this section, we provide more detailed explanations regarding multi-exit architectures as well as curriculum learning, which are the foundations of our method. We start by describing the mathematical model for multi-exit architectures and listing popular training strategies proposed for such architectures. Subsequently, we elaborate on the curriculum learning strategy, including the concepts of sorting and pacing functions, and recount various approaches to these functions that exist in the literature.

\subsection{Multi-Exit Architectures}
Following the notation of Scardapane et al. \cite{ee_survey}, basic neural networks are formulated as function $ f(x) = f_L(f_{L - 1}(...f_1(x))) $ where $L$ is the number of layers and $f_i$ denotes the operator at layer $i$, which can be a convolution layer, a dense layer, batch normalization or any other differentiable operator. The output of the $i$-th layer is denoted by $h_i = f_i(h_{i - 1})$ where $h_0 = x$, and $\theta_i$ signifies all trainable parameters of layer $i$.

\begin{figure}[htbp]
\centerline{\includegraphics[width=0.48\textwidth]{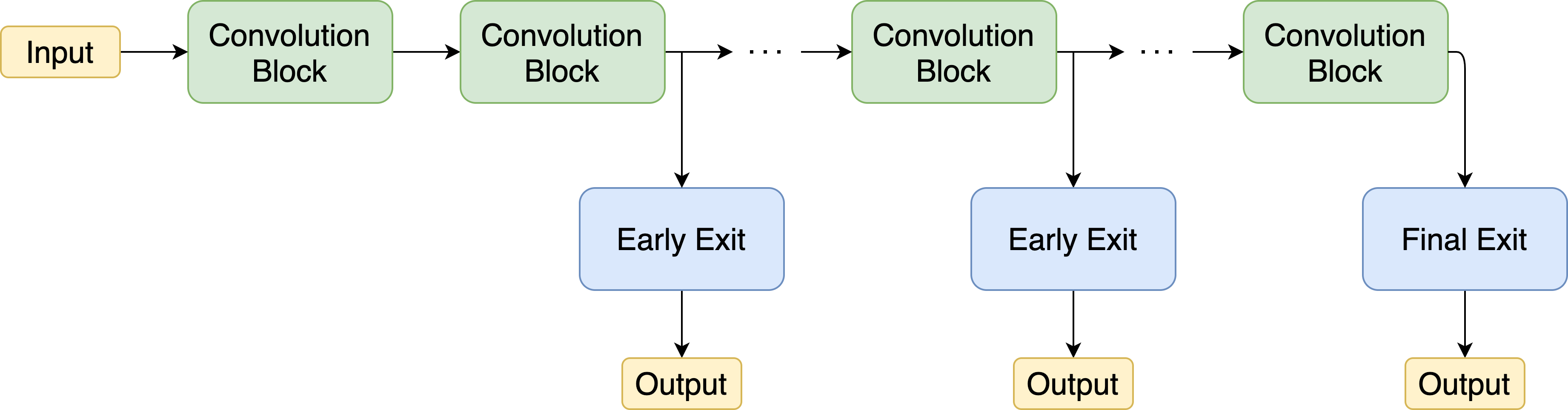}}
\caption{Schematic illustration of a multi-exit architecture.}
\label{multi_exit}
\end{figure}

In order to extend this framework to multi-exit architectures, first, a set of branch locations $B \subseteq \{1, .., L\}$ are selected. For each branch location $b$, a classifier or regressor $c_b(h_b) = y_b $ is defined, where $y_b$ is the hypothesis of the early exit branch at location $b$. The schematic illustration of a multi-exit architecture is depicted in Figure \ref{multi_exit}.

The training of a neural network can be formulated as tuning its parameters by applying an optimization algorithm on a loss landscape:
\begin{equation}
f^* = \argmin_{\theta} \sum_{n = 1}^N l(y_n, f(x_n)),
\label{basic_nn_loss}
\end{equation}
where $\theta = \bigcup_{i = 1}^L \theta_i$ is the set of all parameters of the neural network, $\{(x_n, y_n)\}_{n = 1}^N$ is the set of training samples, and $l(\cdot)$ is a loss function.

However, due to the attached early exit branches, the training of multi-exit architectures is not as straightforward. Three main approaches were proposed for training a multi-exit architecture \cite{ee_survey, ee_survey2}:
\begin{itemize}
  \item \textit{End-to-End Training:} Training is formulated as a single optimization problem where the total loss is defined as a combination of the losses of early exit branches and the final layer. In this case, the contribution of each of the early exit branches to the total loss is expressed with a weight value (a hyper-parameter) that causes trade-offs and can have a significant impact on the accuracy of the early exit branches as well as the final layer. For instance, a certain weighting scheme for the contribution of branches may result in an increase in the accuracy of early exit branches but a decrease in the accuracy of the final layer.
  
  \item \textit{Layer-Wise Training:} Initially, the entire network up to and including the first early exit branch is trained. Subsequently, the trained weights are frozen, meaning that they are not allowed to be modified anymore, and the rest of the network up to and including the second early exit branch is trained. This operation is repeated until the entire network has been trained. Note that with this strategy, there is no guarantee that the accuracy of the final layer will be similar to the case where the network does not have any early exit branches.
  
  \item \textit{Classifier-Wise Training:} The entire backbone network is initially trained. Then, the parameters of the backbone network are frozen and each branch is trained separately since it does not affect the training of other early exit branches. Note that no trade-offs are introduced in this strategy, and since the parameters of the backbone network are not modified, its accuracy remains unchanged. However, the early exit branches have less parameters available for training compared to the other two strategies.
\end{itemize}

In this work we follow the classifier-wise training strategy for training the multi-exit architectures because of its practical importance. This is due to the fact that it can be easily added on top of existing networks (as a ``plug-and-play'' solution) without the need for re-training a high-performing backbone network, or computationally expensive and tedious experimentation for determining the optimal hyper-parameters that lower the effect of trade-offs introduced by combined training of the parameters of the early exit branches with those of the backbone network. 
Furthermore, one of the issues with multi-exit architecture is choosing the right number of early exit branches and their placement. With end-to-end and layer-wise training strategies, the choice of the total number of branches as well as their placement in the backbone network becomes important and can cause further trade-offs. On the other hand, with the classifier-wise training strategy, since the branches are independent of each other and the backbone network, early exit branches can be placed at any intermediate layer. However, we need to keep in mind that early exit branches placed later in the network do not necessarily achieve a higher accuracy, therefore some branch placements may be irrational and unnecessary since there are earlier branches which can potentially achieve higher accuracy.

Another concern with multi-exit architectures is devising a method that decides which exit should be used for each input example. As previously mentioned, a simple solution is to use the confidence of the network on its own prediction, although many other methods have been proposed for this purpose \cite{ee_survey}. However, since our goal is to develop a method in order to increase the accuracy of all early exit branches regardless of their placement, this issue is outside the scope of this paper.

\subsection{Curriculum Learning}
As previously stated, curriculum learning draws inspiration from the way humans learn new subjects throughout their formal education. For each topic of study, a knowledgeable teacher often starts with explaining the simplest notions to the students and gradually introduces more difficult aspects of the topic during the course of the study. Curriculum learning treats the problem of training neural networks in the same manner by starting the training from a subset of training samples it deems to be simple, and progressively adding more difficult samples to the training process. Thus, curriculum learning is composed of two main components: a \textit{sorting function} that takes training samples as input, assigns a difficulty value to each of them based on some metric and sorts them based on their difficulty values; and a \textit{pacing function} that determines the pace at which new training samples are introduced to the network during the training process.

Scoring functions can either be predefined, meaning that the difficulty for each training sample is determined based on some prior knowledge given by an expert, or automatic, meaning that the difficulty of each sample is determined based on an algorithm. 
Examples of predefined sorting functions include sorting based on the length of the input text in natural language processing problems, or based on the number of objects in an image in object detection problems. A comprehensive list of predefined sorting functions for various types of data can be found in \cite{cl_survey}.

Most automatic sorting functions can be categorized into the following three groups \cite{cl_survey}:
\begin{itemize}
    \item \textit{Self-Paced Learning:} In this approach, the student network itself determines the difficulty of each sample based on its current loss. It is important to note that Hacohen et al. \cite{hacohen} found that self-paced learning can lead to a decrease in the final test accuracy.
    
    \item \textit{Transfer Teacher:} In this strategy, the loss of a pre-trained network called \textit{teacher} is used to measure the difficulty of training samples. A variant of transfer teacher where the teacher network is the same as the backbone network is called \textit{self-taught} (not to be confused with \textit{self-paced learning}). The main difference between self-taught and other teacher transfer methods is that the self-taught method can be applied repeatedly, meaning that initially the network is trained normally and its losses are used to sort the examples and train the same network with curriculum learning. Afterwards, the losses of the new and improved network are used to re-sort the training samples and train the same network yet another time, and this process can be repeated until there are no further improvements.
    
    \item \textit{Reinforcement Learning Teacher:} Curriculum learning can also be formulated as a reinforcement learning problem where the action is to decide which samples should be used for training, the state is the loss of the student for each sample, and the reward is the performance of the student.
\end{itemize}

Several other less common automatic sorting functions can be found in \cite{cl_survey}. In this work, we use the \textit{transfer teacher} method with two different teacher networks as scoring function. As previously mentioned, unlike human learning, the opposite approach of training the network starting from the most difficult samples to the easiest samples, called \textit{anti-curriculum} or \textit{harder-first}, has been shown to be more effective than curriculum learning in some cases \cite{cl_survey}.

Typically, a pacing function $ \lambda(t) : \mathbb{N} \rightarrow (0, 1] $ takes the index of the current iteration as an input and outputs the fraction of the sorted training samples that should be used for training. Pacing functions can be categorized into two groups: discrete pacing functions and continuous pacing functions. The most popular discrete pacing function, called \textit{baby step}, partitions sorted training samples into several buckets and gradually adds buckets of harder samples to the pool of training samples introduced to the network. A less common discrete pacing function called \textit{one-pass} partitions the sorted training samples into several buckets, but discards the the samples of the previously introduced easier bucket from the training pool after adding the samples of a new harder bucket.

Popular examples of continuous pacing functions include \textit{linear}, \textit{root}, \textit{root-p} and \textit{geometric progression}, which are described by Equations (\ref{linear})-(\ref{geom}) respectively:
\begin{equation}
\lambda_{\mathrm{linear}}(t) = \min \left(1, \lambda_0 + \frac{1 - \lambda_0}{T_f} \cdot t \right),
\label{linear}
\end{equation}
\begin{equation}
\lambda_{\mathrm{root}}(t) = \min \left(1, \sqrt{\lambda_0^2 + \frac{1 - \lambda_0^2}{T_f} \cdot t} \right),
\label{root}
\end{equation}
\begin{equation}
\lambda_{\mathrm{root{\text{-}}p}}(t) = \min \left(1, \sqrt{\lambda_0^p + \frac{1 - \lambda_0^p}{T_f} \cdot t} \right),
\label{root_p}
\end{equation}
\begin{equation}
\lambda_{\mathrm{geom}}(t) = \min \left(1, 2^{(\log_2 \lambda_0 - \frac{\log_2 \lambda_0}{T_f} \cdot t)} \right).
\label{geom}
\end{equation}
In the above equations, $t$ is the index of current iteration, $\lambda_0$ denotes the initial fraction of training samples introduced to the network and $ T_f $ is the iteration at which the entire dataset is used for the first time.

\begin{figure}[htbp]
\centerline{\includegraphics[width=0.48\textwidth]{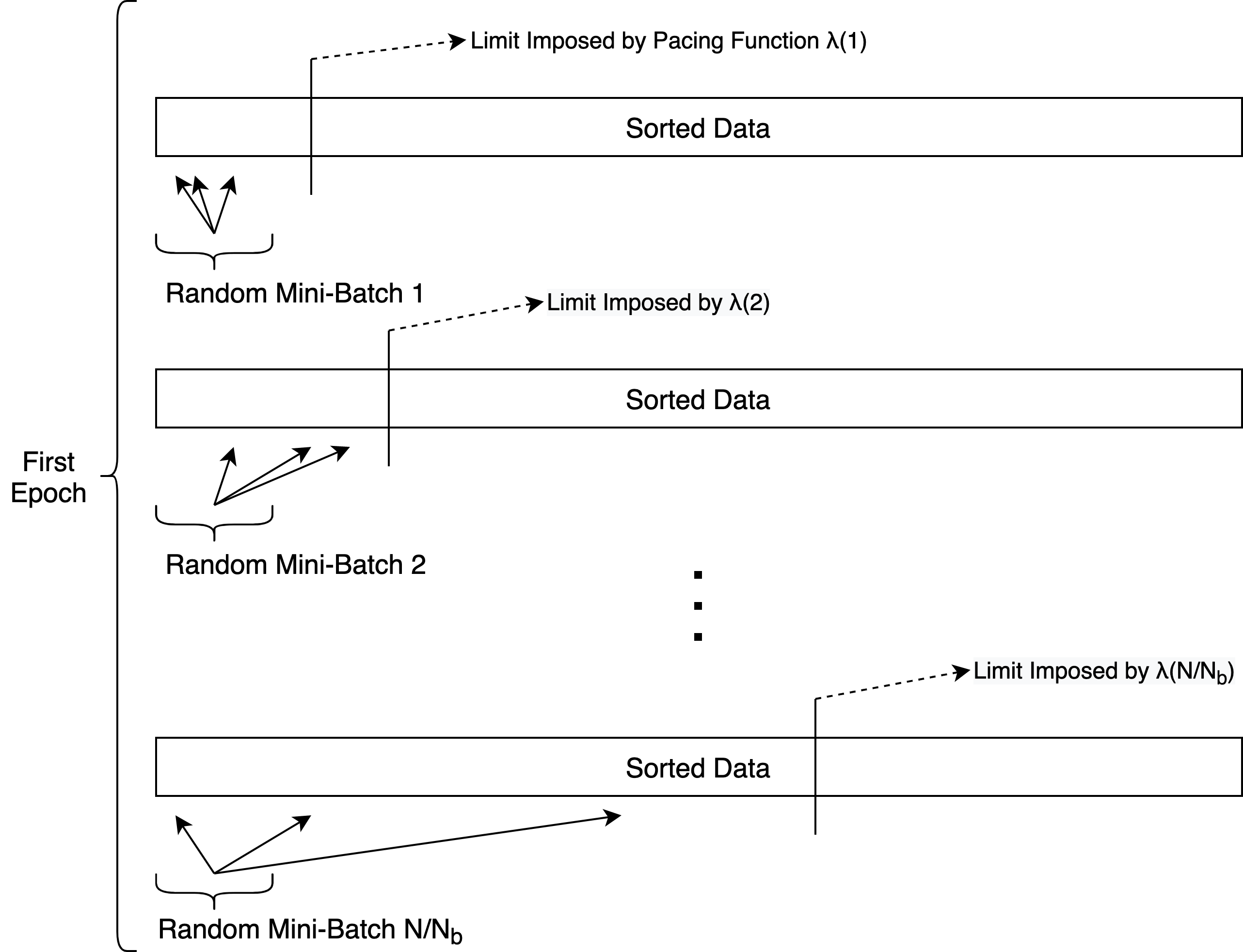}}
\caption{Random Mini-Batch Process in Curriculum Learning.}
\label{cl_mini_batch}
\end{figure}

Putting it all together, Figure \ref{cl_mini_batch} shows the random mini-batch process in curriculum learning. Each epoch is composed of $\frac{N}{N_b}$ batches where $N$ is the total number of training samples and $N_b$ is the batch size. Batch number $t$ is sampled uniformly at random only from the first $\lambda(t)$ portion of the sorted data.

As a final note, there are several theoretical analyses in the literature explaining why curriculum learning can improve the training procedure. Bengio et al. \cite{original_cl} point out that curriculum learning can be viewed as a \textit{continuation method}. Continuation methods \cite{continuation_methods} are optimization strategies for non-convex problems that start with a smooth objective and gradually introduce less smooth versions in the hopes of revealing the global picture in the process \cite{cl_survey}. Additionally, Hacohen et al. \cite{hacohen} reached the conclusion that curriculum learning modifies the optimization landscape to amplify the difference between the optimal parameter vector and all other vectors that have a small covariance with the optimal solution, including uncorrelated or negatively correlated parameter vectors.

\section{Multi-Exit Curriculum Learning}\label{S:Methods}
In this section, we will explain the details of our method. We assume that an already trained high-performing deep neural network is given in the beginning. Due to time restrictions, this neural network must be converted to a multi-exit architecture, as it is preferable to provide an output within the strict time budget, even though it can be less accurate, rather than not providing an output within this time limit at all. Thus we augment this backbone network with a set of early exits. As previously stated, the parameters of the backbone network will not be fine-tuned, that is, if the backbone network represents function $f(x) = f_L(...f_1(x))$ with a set of parameters $\theta = \bigcup_{i = 1}^L \theta_i $, $\theta$ will remain unchanged throughout the training process and only the parameters of early exit branch functions $c_i(h_i) : i \in B$ will be tuned. 
As the entire backbone network is frozen during classifier-wise training of the added early exit branches, and thus is not allowed to ``help'' the early exit branches by tuning its parameters, it is more difficult to increase the accuracy of the early exit branches compared to other training strategies listed in Section \ref{S:RelatedWork}. We use curriculum learning to train the early exit branches, in order to improve their accuracy.

For the purpose of sorting the training samples based on their difficulty, we use the categorical cross-entropy loss of a pre-trained teacher network. We use two different teachers, InceptionV3 \cite{inception_v3} which is the same teacher used in Hacohen et al. \cite{hacohen}, and the more recent EfficientNetB7 \cite{efficientnet}. We take versions of these networks pre-trained on the ImageNet dataset \cite{imagenet} and use transfer learning to train them for the CIFAR-10 and CIFAR-100 datasets by removing the top layer, adding two dense layers with a Dropout layer \cite{dropout} in between and retraining the network for the intended dataset. By using two dense layers, we are taking the output of pre-trained networks as feature vectors and training a multilayer perceptron classifier based on these features. In addition, since we freeze the first five blocks of the EfficientNetB7 backbone to overcome the limitations of our hardware resources, utilizing two dense layers instead of just one provides additional flexibility.

\begin{figure}[htbp]
\centerline{\includegraphics[width=0.48\textwidth]{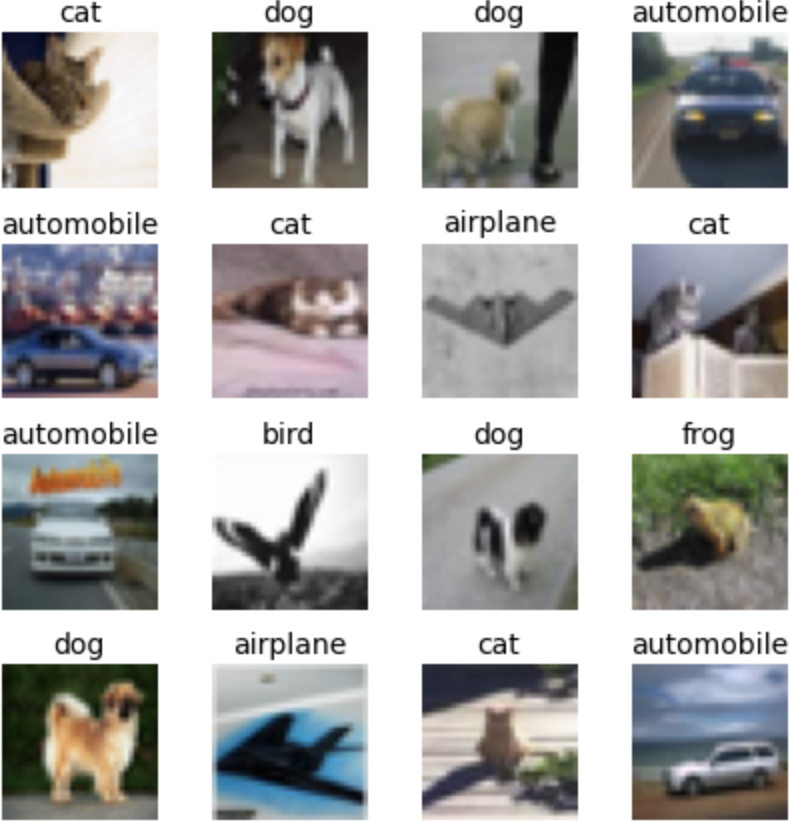}}
\caption{Easiest training samples in the CIFAR-10 dataset based on the loss values of InceptionV3 teacher network. The labels are ground truth, not the predictions of the network.}
\label{easiest}
\end{figure}

\begin{figure}[htbp]
\centerline{\includegraphics[width=0.48\textwidth]{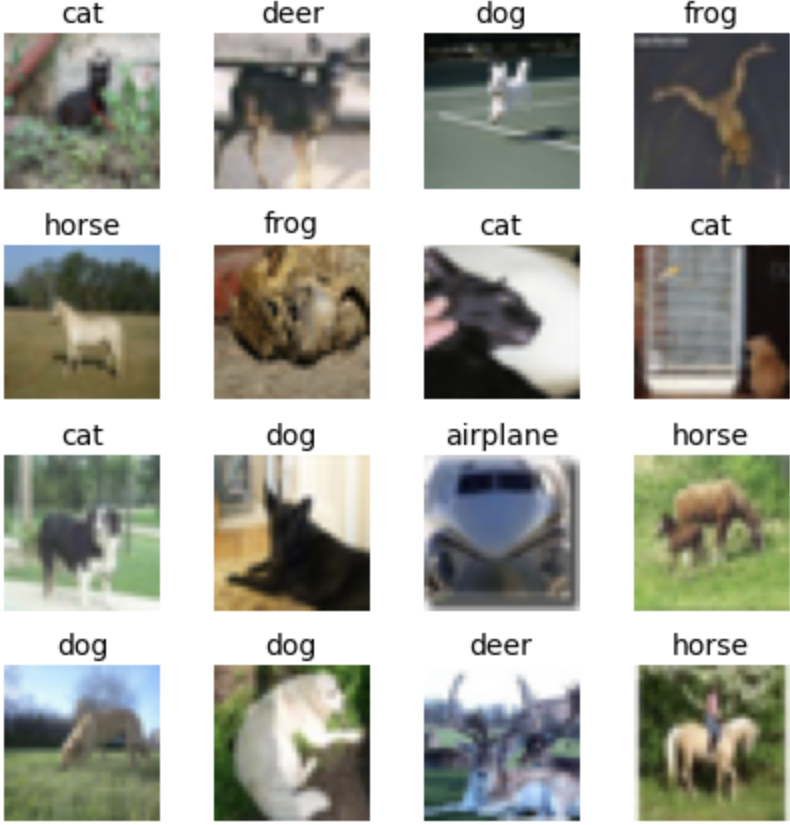}}
\caption{Hardest training samples in the CIFAR-10 dataset based on the loss values of InceptionV3 teacher network. The labels are ground truth, not the predictions of the network.}
\label{hardest}
\end{figure}

Figures \ref{easiest} and \ref{hardest} illustrate the easiest and most difficult training samples, respectively, in the CIFAR-10 dataset based on the loss values of InceptionV3 teacher. It is not difficult to interpret why the network finds some of these images  particularly hard. For instance, a close-up from the front of the airplane might be very different from the usual perspective of other images with the same label, or it may be difficult to distinguish between dogs, cats and deer with certain colors and patterns of fur.

\begin{figure}[htbp]
\centerline{\includegraphics[width=0.48\textwidth]{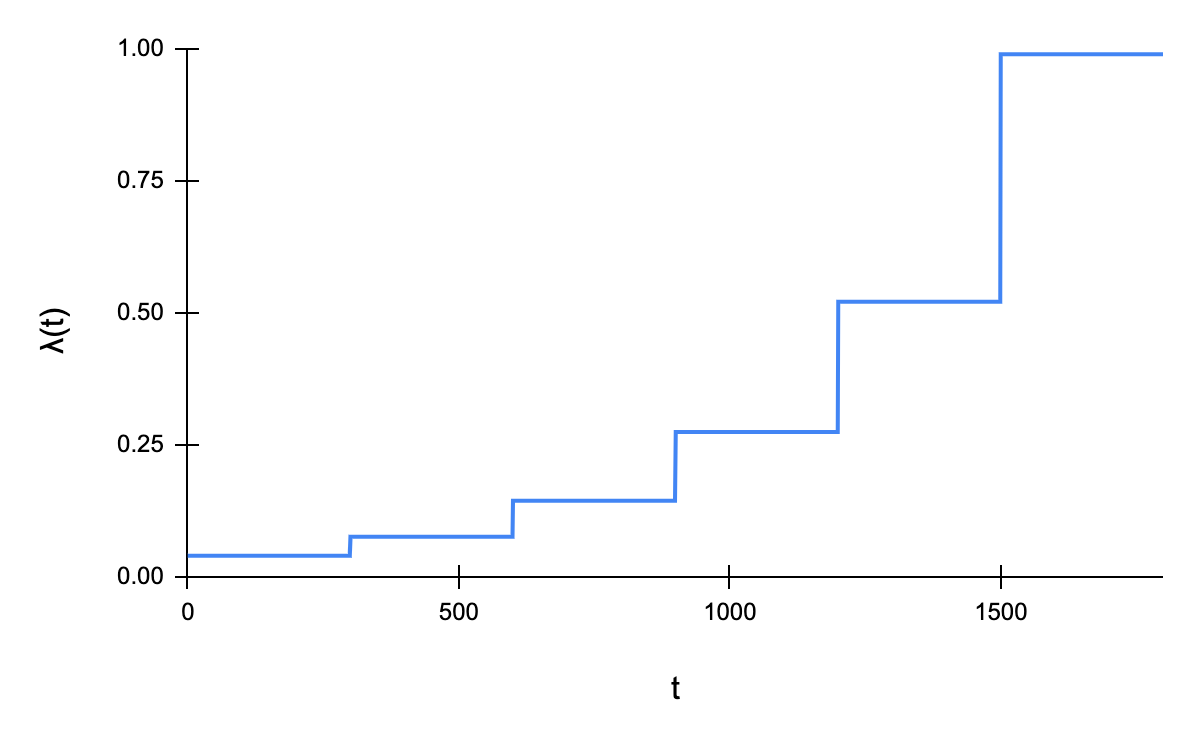}}
\caption{\textit{Fixed exponential pacing} function with $ s = 0.04 $, $ r = 1.9 $ and $ \delta = 300 $.}
\label{fep}
\end{figure}

\begin{figure}[htbp]
\centerline{\includegraphics[width=0.48\textwidth]{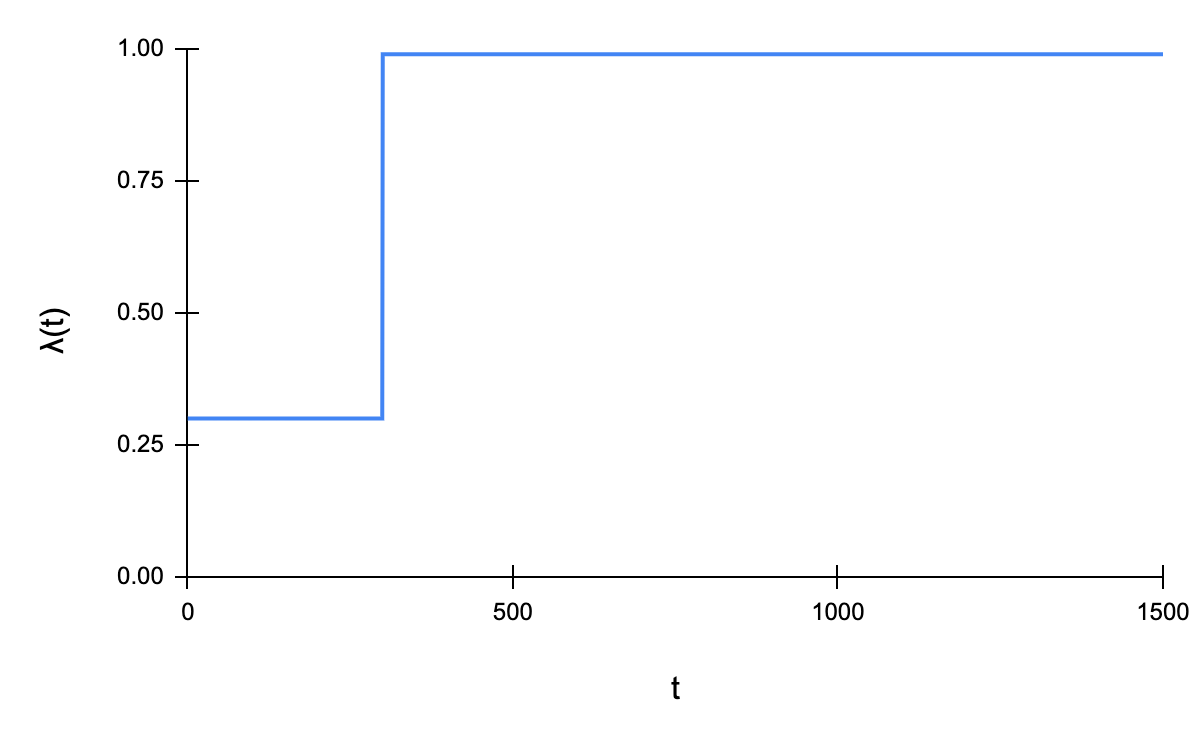}}
\caption{\textit{Single step pacing} function with $ s = 0.30 $ and $ \delta = 300 $.}
\label{ssp}
\end{figure}

We use two variants of the \textit{baby step} pacing function, the \textit{fixed exponential pacing} function shown in Fig. \ref{fep} and the \textit{single step pacing} function depicted in Fig. \ref{ssp}, both introduced by Hacohen et al. \cite{hacohen}. Similar to our work, Hacohen et al. \cite{hacohen} also investigate the effectiveness of curriculum learning on the problem of image classification (although not in multi-exit architectures) and document the pacing functions that lead to improvements in the final accuracy. These pacing functions introduce the entire dataset fairly quickly, meaning that curriculum learning effectively takes place only in the first few epochs. We found that such pacing functions are effective in our case as well. \textit{Fixed exponential pacing} starts with only a small percentage of the training data and exponentially increases the amount of data after every fixed number of batches, whereas \textit{single step pacing} starts with a higher percentage of data and introduces the entire dataset after a certain number of batches have been processed. The details of \textit{fixed exponential pacing} and \textit{single step pacing} functions are shown in Equations \eqref{fep_eq} and \eqref{ssp_eq} respectively, where $ t $ is the index of the current batch, $ s $ indicates the initial fraction of data used, $ r $ denotes the increase in data and $ \delta $ is the fixed number of batches after which the data is increased. It is important to note that \textit{fixed exponential pacing} has three hyper-parameters, namely $ s $, $ r $ and $ \delta $, while \textit{single step pacing} has only two.

\begin{equation}
\lambda(t) = \min \left(\mathit{s} \cdot \mathit{r}^{\lfloor\frac{t}{\mathit{\delta}}\rfloor}, 1 \right)
\label{fep_eq}
\end{equation}

\begin{equation}
\lambda(t) = \left\{ \begin{array}{cc} 
    \mathit{s}, & \hspace{5mm} t < \mathit{\delta} \\
    1, & \hspace{5mm} t \geq \mathit{\delta} \\
    \end{array} \right.
\label{ssp_eq}
\end{equation}

We use four different backbone networks in our experiments, namely DenseNet201 \cite{densenet}, MobileNetV1 \cite{mobilenet}, ResNet152 \cite{resnet} and InceptionV3 \cite{inception_v3}. We train these networks on the CIFAR-10 and CIFAR-100 datasets using transfer learning in the exact same way as the aforementioned teacher networks. 

In the training of teacher and backbone networks, in order to overcome the limitations of our available resources, the size of the batches are adjusted and some of the layers in the networks are frozen, that is, their weights are not modified during the training process. Keep in mind that since these networks are all pre-trained on the ImageNet dataset, the frozen layers are still capable of providing useful features. Table \ref{training_details} summarizes the details of the training process for each of these networks.

\begin{table}[htbp]
\caption{Training Details for Teacher and Backbone Networks}
\begin{center}
\resizebox{\linewidth}{!}{
\begin{tabular}{ p{2.0cm} p{1.0cm} p{2.0cm} p{1.0cm} p{1.0cm} }
\hline
Network & Batch Size & Frozen Layers & \multicolumn{2}{c}{Test Accuracy}\\
& & & CIFAR-10 & CIFAR-100\\
\hline
\hline
DenseNet201 & 32 & All except batch normalization & 96.48\% & 82.53\%\\
\hline
MobileNetV1 & 64 & None & 94.28\% & 76.91\%\\
\hline
ResNet152 & 32 & All except batch normalization & 95.36\% & 82.25\&\\
\hline
InceptionV3 & 64 & None & 96.56\% & 83.80\%\\
\hline
EfficientNetB7 & 32 & First five blocks & 96.50\% & 83.76\%\\
\hline
\end{tabular}
}
\end{center}
\label{training_details}
\end{table}

\begin{figure}[htbp]
\centerline{\includegraphics[width=0.48\textwidth]{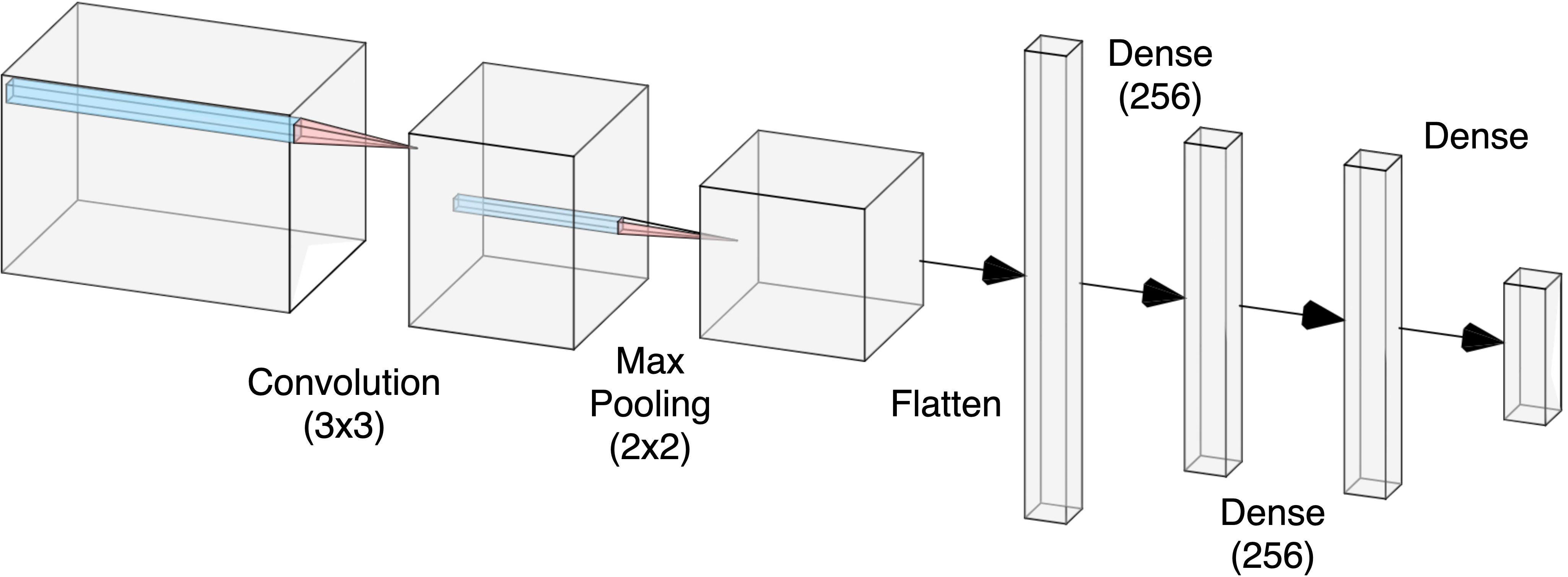}}
\caption{Architecture of Early Exit Branches.\protect\footnotemark}
\label{branch_arch}
\end{figure}

We place two early exit branches at two different intermediate layers on each backbone network. All early exit branches have the same architecture, which is a convolution layer, followed by a maximum pooling layer, and three dense layers with a Dropout layer between each pair, as shown in Figure \ref{branch_arch}. Note that since the dimention of features in different branch locations might be different, the size of the flattened vector varies for each branch location. This is the same branch architecture used by by Hu et al. \cite{triple_wins}. The location of each branch depends on the architecture of the backbone network. We found that placing an early exit branch later in the backbone network does not necessarily improve the overall accuracy of the branch, and generally speaking branches located immediately after the ``natural blocks'' - for instance, concatenation layers, residual connections or dense blocks - in the architecture performed better than several other layers immediately before or after them. Early exit branches are placed at the earlier sections of the backbone network as they correspond to the locations where dynamic inference would be desired in practical scenarios. The exact placement of branches for each backbone network can be found in Tab. \ref{exit_locations}.

\footnotetext{Image created using the NN-SVG tool\cite{nn_svg}.}

\begin{table}[htbp]
\caption{Placement of Branches for Each Backbone Network}
\begin{center}
\resizebox{\linewidth}{!}{
\begin{tabular}{ c c c | c } 
\hline
    Backbone &
    Dataset &
    BN$^{*}$ &
    Branch Placed After \\
\hline
\hline
    DenseNet201 &
    CIFAR-10 &
    1 &
    Layer 15 of 201 \\
    
    &
    &
    2 &
    Layer 40 of 201 \\

    &
    CIFAR-100 &
    1 &
    Layer 40 of 201 \\

    &
    &
    2 &
    Layer 137 of 201 \\

\hline

    MobileNet &
    CIFAR-10 &
    1 &
    Layer 8 of 28 \\

    &
    &
    2 &
    Layer 14 of 28 \\

    &
    CIFAR-100 &
    1 &
    Layer 8 of 28 \\

    &
    &
    2 &
    Layer 14 of 28 \\

\hline

    ResNet152 &
    CIFAR-10 &
    1 &
    Layer 13 of 152 \\

    &
    &
    2 &
    Layer 38 of 152 \\

    &
    CIFAR-100 &
    1 &
    Layer 13 of 152 \\

    &
    &
    2 &
    Layer 38 of 152 \\

\hline

    InceptionV3 &
    CIFAR-10 &
    1 &
    1st Filter Concat \\

    &
    &
    2 &
    2nd Filter Concat \\

    &
    CIFAR-100 &
    1 &
    1st Filter Concat \\

    &
    &
    2 &
    2nd Filter Concat \\
\hline
\multicolumn{4}{l}{$^{*}$Branch Number} \\
\end{tabular}
}
\end{center}
\label{exit_locations}
\end{table}

As previously mentioned, we use the classifier-wise training strategy for training the multi-exit architecture. During the training of each branch, first we test both stochastic gradient descent and Adam \cite{adam} optimizers with different learning rates of $ \{10^{-1}, 0.12, 10^{-2}, 10^{-3}, 10^{-4}, 10^{-5} \} $ to obtain the highest accuracy for the normal training method without any curriculum, which we call \textit{vanilla}. We chose to test the $ 0.12 $ learning rate in addition to $ 10^{-1} $ since it was the best case discovered for the experiments in Hacohen et al. \cite{hacohen}. With both optimizers, the learning rate is automatically reduced when the validation accuracy plateaus. Subsequently, using the same optimizer, we train the branch using the \textit{curriculum} and \textit{anti-curriculum} training methods. Similar to Hacohen et al. \cite{hacohen}, we also compare the results with \textit{random curriculum} which uses the pacing function on randomly-ordered training data without any sorting. This comparison is in order to show that the benefit does not solely come from the pacing, and that the sorting from easiest to hardest or vice-versa contributes to the increase in accuracy as well.

As with many machine learning paradigms, curriculum learning is sensitive to hyper-parameters, therefore we perform a grid search in order to find the suitable teacher network and pacing function in each case. We use a dataset separate from training, validation and test datasets for this task. Table \ref{hyper_parameter} summarizes the different choices of pacing functions tested during the hyper-parameter optimization step.

\begin{table}[htbp]
\caption{Hyper-parameter Optimization for Pacing Function}
\begin{center}
\begin{tabular}{ l c c c c } 
\hline
Function Type & \multicolumn{3}{c}{Parameters} & Abbreviation \\
& $ \mathit{s} $ & $ \mathit{r} $ & $ \mathit{\delta} $ \\
\hline
\hline
Fixed Exponential Pacing & 0.04 & 1.9 & 100 & FEP(100)\\
Fixed Exponential Pacing & 0.04 & 1.9 & 200 & FEP(200) \\
Fixed Exponential Pacing & 0.04 & 1.9 & 300 & FEP(300) \\
Single Step Pacing & 0.30 & - & 300 & SSP(300) \\
\hline
\end{tabular}
\end{center}
\label{hyper_parameter}
\end{table}

\begin{table*}[htbp]
\caption{Comparison of the Final Test Accuracy of Early Exit Branches using Different Training Methods}
\begin{center}
\resizebox{\linewidth}{!}{
\begin{tabular}{ c c c | c c c c | c c c c } 
\hline
    Backbone &
    Dataset &
    BN$^{*}$ &
    Vanilla &
    Curriculum &
    AC$^{\dagger}$ &
    RC$^{\mathsection}$ &
    Opt. &
    LR$^{\mathparagraph}$ &
    Teacher &
    Pacing \\
\hline
\hline
    DenseNet &
    CIFAR-10 &
    1 &
    71.71\% ± 0.47 &
    71.59\% ± 0.57 &
    \textbf{71.75\% ± 0.75} &
    71.58\% ± 0.57 &
    Adam &
    $ 10^{-4} $ &
    EfficientNet &
    FEP(100) \\
    
    &
    &
    2 &
    77.43\% ± 0.64 &
    \textbf{77.91\% ± 0.03} &
    77.21\% ± 0.80 &
    77.25\% ± 0.35 &
    SGD &
    $ 0.12 $ &
    EfficientNet &
    FEP(100) \\

    &
    CIFAR-100 &
    1 &
    38.36\% ± 0.31 &
    \textbf{39.56\% ± 0.57} &
    35.32\% ± 2.06 &
    38.74\% ± 1.37 &
    SGD &
    $ 0.12 $ &
    EfficientNet &
    FEP(200) \\

    &
    &
    2 &
    61.72\% ± 1.26 &
    \textbf{64.05\% ± 1.18} &
    58.78\% ± 1.68 &
    62.95\% ± 0.78 &
    Adam &
    $ 10^{-4} $ &
    EfficientNet &
    FEP(300) \\

\hline

    MobileNet &
    CIFAR-10 &
    1 &
    67.30\% ± 0.25 &
    \textbf{67.33\% ± 0.31} &
    67.04\% ± 0.45 &
    67.02\% ± 0.48 &
    Adam &
    $ 10^{-4} $ &
    EfficientNet &
    SSP(300) \\

    &
    &
    2 &
    79.06\% ± 0.65 &
    \textbf{79.47\% ± 0.05} &
    79.04\% ± 0.43 &
    78.55\% ± 0.41 &
    Adam &
    $ 10^{-4} $ &
    Inception &
    FEP(100) \\

    &
    CIFAR-100 &
    1 &
    44.26\% ± 0.69 &
    44.83\% ± 0.19 &
    \textbf{44.89\% ± 0.26} &
    44.84\% ± 0.45 &
    Adam &
    $ 10^{-4} $ &
    Inception &
    FEP(300) \\

    &
    &
    2 &
    47.48\% ± 0.99 &
    \textbf{48.39\% ± 0.66} &
    47.54\% ± 0.45 &
    48.34\% ± 0.74 &
    Adam &
    $ 10^{-4} $ &
    EfficientNet &
    FEP(200) \\

\hline

    ResNet &
    CIFAR-10 &
    1 &
    67.87\% ± 0.76 &
    \textbf{68.75\% ± 0.16} &
    67.78\% ± 0.26 &
    67.44\% ± 0.74 &
    Adam &
    $ 10^{-4} $ &
    EfficientNet &
    FEP(100) \\

    &
    &
    2 &
    76.25\% ± 0.25 &
    76.24\% ± 0.28 &
    \textbf{76.32\% ± 0.25} &
    76.29\% ± 0.11 &
    Adam &
    $ 10^{-4} $ &
    EfficientNet &
    FEP(100) \\

    &
    CIFAR-100 &
    1 &
    35.53\% ± 0.74 &
    \textbf{36.57\% ± 1.07} &
    36.46\% ± 0.82 &
    36.08\% ± 0.70 &
    Adam &
    $ 10^{-4} $ &
    EfficientNet &
    SSP(300) \\

    &
    &
    2 &
    41.26\% ± 0.56 &
    41.30\% ± 1.02 &
    \textbf{41.45\% ± 0.73} &
    40.89\% ± 0.36 &
    Adam &
    $ 10^{-4} $ &
    EfficientNet &
    FEP(100) \\

\hline

    Inception &
    CIFAR-10 &
    1 &
    76.91\% ± 0.58 &
    \textbf{77.34\% ± 0.27} &
    77.13\% ± 0.07 &
    77.19\% ± 0.17 &
    Adam &
    $ 10^{-4} $ &
    EfficientNet &
    FEP(300) \\

    &
    &
    2 &
    79.06\% ± 0.37 &
    79.18\% ± 0.12 &
    \textbf{79.47\% ± 0.52} &
    79.42\% ± 0.15 &
    Adam &
    $ 10^{-4} $ &
    Inception &
    FEP(100) \\

    &
    CIFAR-100 &
    1 &
    44.24\% ± 0.70 &
    \textbf{44.56\% ± 0.44} &
    44.53\% ± 0.42 &
    44.07\% ± 0.75 &
    Adam &
    $ 10^{-4} $ &
    Inception &
    FEP(200) \\

    &
    &
    2 &
    45.86\% ± 0.21 &
    \textbf{46.50\% ± 0.21} &
    45.13\% ± 0.92 &
    46.11\% ± 1.28 &
    Adam &
    $ 10^{-4} $ &
    Inception &
    FEP(300) \\
\hline
\multicolumn{10}{l}{$^{*}$Branch Number} \\
\multicolumn{10}{l}{$^{\dagger}$Anti-Curriculum} \\
\multicolumn{10}{l}{$^{\mathsection}$Random Curriculum} \\
\multicolumn{10}{l}{$^{\mathparagraph}$Learning Rate}
\end{tabular}
}
\end{center}
\label{accuracies}
\end{table*}

We repeat each of the experiments five times and record the average accuracy along with the standard deviation. In order to make the comparisons fair, in each repetition, in all four cases of \textit{vanilla}, \textit{curriculum}, \textit{anti-curriculum} and \textit{random curriculum}, the early exit branch starts with the same weight initialization.

\section{Results}\label{S:Results}
Our results are summarized in Table \ref{accuracies}. The first three columns of the table determine the case under study; the next four columns compare the final accuracy of different training approaches for each case; and the last four columns summarize the optimal hyper-parameters discovered for each case. We can observe that in all 16 cases, the accuracy of our method (curriculum and anti-curriculum) is higher than the accuracy of the models trained following the vanilla approach. Notice that in five of the cases the accuracy of the anti-curriculum strategy is better than that of curriculum. The fact that anti-curriculum can achieve superior results is hardly surprising, since there are many documented cases in the literature where the anti-curriculum approach yields higher performance than the curriculum approach \cite{cl_survey}. One possible explanation is that anti-curriculum forces the network to focus on the boundary cases and ambiguous examples early on and thus performs better when separating the classes. We can also observe that for three of the cases involving the Inception backbone network, the selected teacher is the Inception network as well. Thus these cases are examples of self-taught teacher transfer. Finally, we note that the best optimizer found for two of the DenseNet cases is SGD while in all other cases the Adam optimizer is selected.

\section{Discussion and Future Directions}\label{S:Conclusions}
In this paper, we proposed a robust way of increasing the accuracy of early exits branches in multi-exit architectures and showed that it works across different datasets, backbone networks, branch locations and optimizers. 
We have also shown that even though curriculum learning provides the highest accuracy in most of the experiments, in some cases, anti-curriculum achieves the best performance, therefore it is better to test both of these approaches rather than relying exclusively on the former. 

As future research directions, it would be worth investigating how our method performs with other training strategies for multi-exit architectures. It is unlikely that this method increases the accuracy for every weighting scheme of the end-to-end training strategy or for every layer in the layer-wise training strategy, however, it is an interesting problem to discover the conditions under which it provides a benefit. Moreover, the combination of our method with other methods for increasing the accuracy of early exits such as the distillation-based training introduced in Phuong et al. \cite{ee_kd} can be explored. Indeed, curriculum learning in combination with an ensemble approach has been shown to improve knowledge distillation \cite{prev_cl}.

\section*{Acknowledgment}
This work was partly funded by the European Union’s Horizon 2020 research and innovation programme under grant agreement No 957337, and by the Danish Council for Independent Research under Grant No. 9131-00119B. This publication reflects the authors views only. The European Commission and the Danish Council for Independent Research are not responsible for any use that may be made of the information it contains.

\end{document}